\documentclass[10pt,twocolumn,letterpaper]{article}

\usepackage{iccv}
\usepackage{times}
\usepackage{epsfig}
\usepackage{graphicx}
\usepackage{amsmath}
\usepackage{amssymb}

\usepackage{kotex}
\usepackage{multirow}
\newcommand{\figref}[1]{Fig.~\ref{#1}}

\newcommand{\tabref}[1]{Table~\ref{#1}}

\newcommand{\secref}[1]{Sec.~\ref{#1}}
\newcommand{\Figref}[1]{Figure~\ref{#1}}
\newcommand{\ourmodel}{PixelHuman}

\usepackage[pagebackref=true,breaklinks=true,letterpaper=true,colorlinks,bookmarks=false]{hyperref}

\iccvfinalcopy 

\def\iccvPaperID{9229} 

\ificcvfinal\fi

\begin{document}

\title{PixelHuman: Animatable Neural Radiance Fields from Few Images}
    

\author{Gyumin Shim \and Jaeseong Lee \and Junha Hyung \and Jaegul Choo\\
Korea Advanced Institute of Science and Technology\\
{\tt\small \{shimgyumin, wintermad1245, sharpeeee, jchoo\}@kaist.ac.kr}
}



\maketitle

\ificcvfinal\thispagestyle{empty}\fi

\begin{abstract}

    In this paper, we propose \ourmodel, a novel human rendering model that generates animatable human scenes from a few images of a person with unseen identity, views, and poses. 
    Previous work have demonstrated reasonable performance in novel view and pose synthesis, but they rely on a large number of images to train and are trained per scene from videos, which requires significant amount of time to produce animatable scenes from unseen human images. 
    Our method differs from existing methods in that it can generalize to any input image for animatable human synthesis. Given a random pose sequence, our method synthesizes each target scene using a neural radiance field that is conditioned on a canonical representation and pose-aware pixel-aligned features, both of which can be obtained through deformation fields learned in a data-driven manner. Our experiments show that our method achieves state-of-the-art performance in multiview and novel pose synthesis from few-shot images.


\end{abstract}

\begin{figure}[t]
\centering
\begin{center}
    \includegraphics[width=1\linewidth]{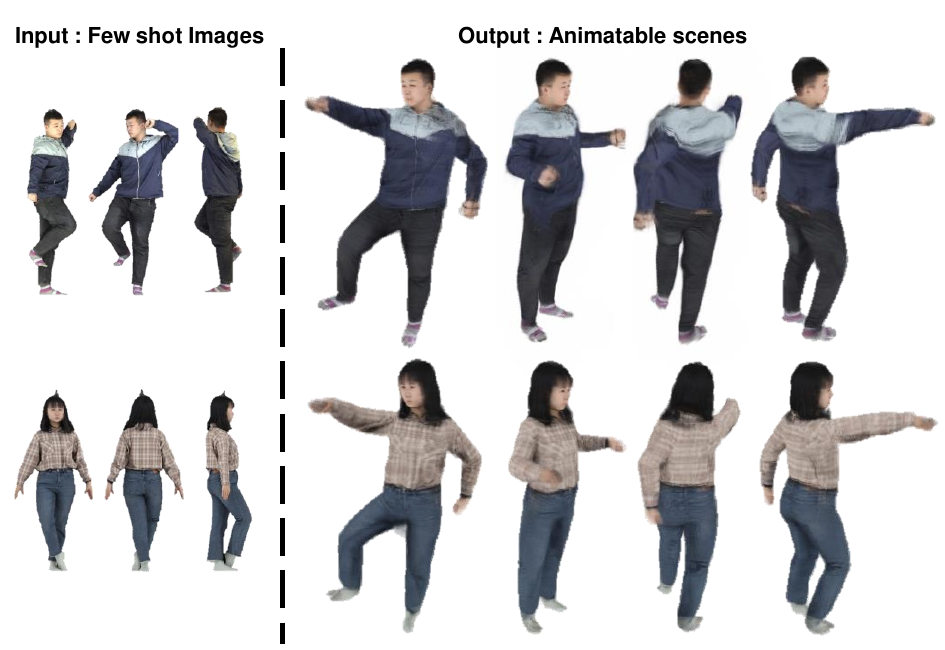}
    \caption{Qualitative results of our method. Given the source images (first column), our method results in realistic posed images from unseen identities given a random pose sequence as input. Note that animation is rendered from the THUman2.0 and Twindom test dataset using the pose sequence in ZJU-MoCap dataset. }
    \label{fig:teaser}
\end{center}
\end{figure}

\section{Introduction}

Reconstructing 3D avatars from humans has a variety of applications of computer vision and graphics, such as virtual reality or metaverse contents. 
Creating human avatars has been developed in various ways, ranging from creating textured human scans using various 3D representations~\cite{saito2019pifu, saito2020pifuhd, zheng2021pamir, shim2022refu, he2020geo}, to rendering human images using implicit representations~\cite{peng2021animatable, weng2022humannerf, li2022tava}.
However, it is an ill-posed problem to reconstruct a 3D human given only a single or few images due to depth ambiguity and occlusions. 
Furthermore, the task becomes even more challenging when attempting to reconstruct 3D contents of a moving person or to animate the person with a random motion sequence. 


As neural radiance fields~\cite{mildenhall2021nerf} has emerged with promising rendering performance on 3D objects, various human rendering methods ~\cite{peng2021animatable, weng2022humannerf, li2022tava} have been proposed to learn 3D human bodies only from images. By learning density and color fields of a 3D human, these methods successfully render human images from novel viewpoint or even in novel pose. 
However, they have limited practicality in many applications as they require thousands of a single or multi-view video frames for optimization. 
Moreover, since they are optimized per-subject, they take a lot of training time if various unseen human models are rendered. 
Although some image-based models~\cite{kwon2021neural, mihajlovic2022keypointnerf} are generalizable to unseen identities, they are limited to multiview synthesis, which does not support novel pose synthesis of given input frames.

Inspired by the recent methods that learn neural radiance fields targeting on human bodies, we propose a novel human rendering model, \textit{\ourmodel}, which generates animatable human scenes from unseen human identity and pose with only a single or few images of the person.
It first learns the skeletal deformation, which is used to map 3D query points in the target space to the different pose spaces. 
To learn the skeletal deformation in a data-driven way, we propose {\it weight field table}, 
which computes unique blend weight fields that are tailored to the body shape of each human identity, enabling a more accurate transformation of 3D query points between different pose spaces.
By learning to reconstruct the exact body shapes of various subjects, the weight field table can learn the latent space of human body shapes, which can be further used to search for new shapes for unseen identities.
Then, our proposed model is able to extract source image features in a pose-aware manner when given the source images that have different poses and views with target spaces.
By incorporating the transformed canonical coordinate and extracted source features, \ourmodel~ successfully renders target images when a random motion sequence is given as an input.  
At test time, we can render animatable scenes of a human body when a few images of the person and a random pose sequence is given as inputs as shown in \Figref{fig:teaser}.

\noindent\ In summary, our contributions are as follows:

\begin{itemize}
\setlength\itemsep{0em}
    \item We propose a novel method that can generate animatable scenes of a moving person conditioned on a single or few images. 
    \item We introduce {\it weight field table} to learn the distinct deformation fields of various objects, which helps reconstruct the exact shape of each human body when rendering various identities. 
    \item The proposed model extracts pixel-aligned features from source images in a pose-aware manner, which allows the model to fully reflect the shape and texture of a person with unseen poses.
    
\end{itemize}

\section{Related Work}

\subsection{Human-targeted NeRF}

As NeRF~\cite{mildenhall2021nerf} has suggested volume rendering technique that is effective for novel view synthesis, numerous studies~\cite{barron2021mip, hedman2021baking,nerf++, nerf--, GRAF, pigan} have been conducted by improving its performance or extending its applications. 
Beyond learning 3D information of static scenes, several extension work~\cite{pumarola2021d, gao2021dynamic, li2021neural} have targeted to generate novel views in videos by learning 3D information of moving scenes. 

Human-targeted NeRFs have been proposed to generate a moving human body by substituting frame components with human pose in dynamic NeRF work.
Some pioneering work~\cite{peng2021neural, weng2020vid2actor} has leveraged volume rendering to render a human body in a motion sequence. 
By learning a set of latent codes structured to SMPL~\cite{loper2015smpl} vertices, NeuralBody~\cite{peng2021neural} successfully renders a human body in a motion sequence given in training videos from any novel viewpoint. 
A-NeRF~\cite{su2021nerf} learns an articulated human representation to render a human body in both unseen views and poses. 
Similarly, Animatable NeRF~\cite{peng2021animatable} is suggested to generate novel human poses by deforming human body into a canonical human model represented by a neural radiance field.
HumanNeRF~\cite{weng2022humannerf} is the recently proposed method that can generate the human body of a moving person by taking a single video as input. 
NeuMan~\cite{jiang2022neuman} further decomposes a video containing a moving person into the background and the human body, enabling scene editing. 
However, all of the stated work have limitations in that a large number of frames are required for training and per-subject optimization is necessary.

\subsection{Few-shot Methods}

Few-shot methods can be categorized into generalizable models and few-shot training models. 
Generalizable models learn prior knowledge from a large-scale dataset in the training stage, and directly predict the output in the inference stage conditioned on the given few-shot images of unseen subjects. 
Few-shot training models, on the other hand, utilize only the given few-shot images to train their models, thus they are required to be optimized independently for every subject. 


Generalization models~\cite{yu2021pixelnerf, wang2021ibrnet} are introduced for few-shot novel view synthesis by predicting neural radiance fields from one or a few images. 
They basically utilize pixel-aligned features by extracting 2D image feature map with a CNN network to directly predict the output in the inference stage.
NHP~\cite{kwon2021neural} introduced a generalizable NeRF model that targets the human body. It synthesizes free-viewpoint images of an unseen moving person when sparse multi-view videos are given as inputs. 
KeypointNeRF~\cite{mihajlovic2022keypointnerf} proposed relative spatial encoding with 3D keypoints, which is proven to be robust to the sparse inputs and cross-dataset domain gap. 
It has achieved the state-of-the-art performance in human head and body reconstruction. 
However, they do not support novel pose synthesis of given input images. 

Few-shot training models~\cite{xu2022sinnerf, dietNeRF} fully utilize other prior knowledge, such as semantic and geometry regularizations, since learning 3D information of objects only from a single or a few images is extremely challenging issue. 
They generally utilize pseudo labels generated from models pre-trained on large-scale datasets.
ELICIT~\cite{huang2022one} targeted to tackle a similar issue in the human body, which is to learn 3D human body from a single image. 
It leverages geometry priors from the SMPL model, semantic priors from CLIP~\cite{radford2021learning} models, and segmentation maps of source human body to learn 3D information in an unsupervised manner. 
However, it still requires per-subject optimization, which limits its practicality in the real world. 

\begin{figure*}[t]
\centering
\begin{tabular}{@{}c}
\includegraphics[width=1\linewidth]{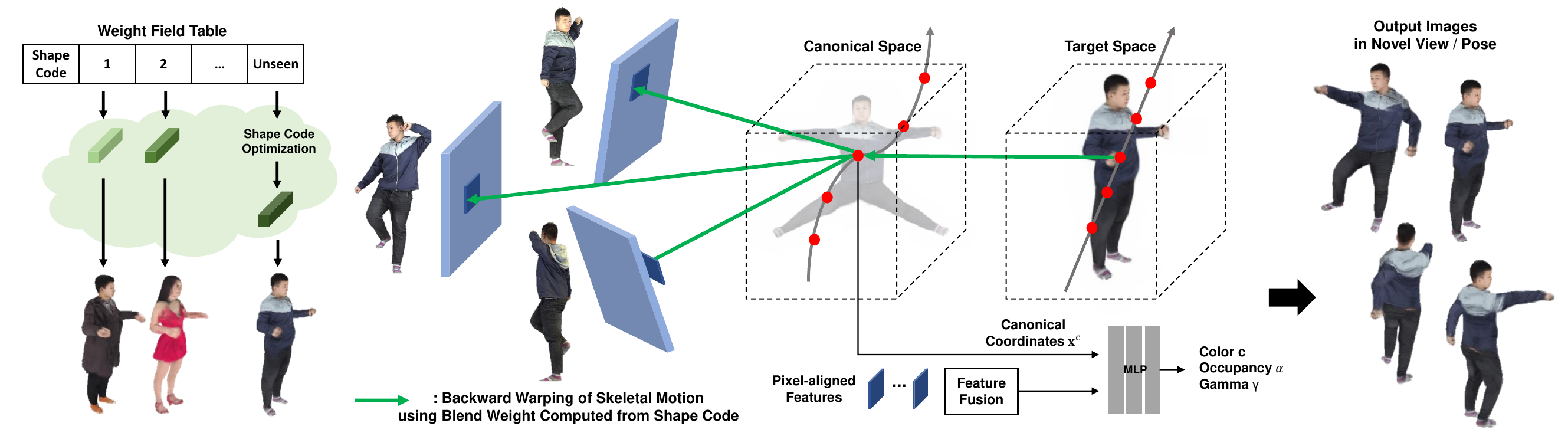} \\
\end{tabular}
\caption{Overview of our approach. Given the target pose in the target space, 3D query points are transformed into the canonical space and the source space to extract the source pixel-aligned feature in a pose-aware manner. Utilizing the transformed canonical coordinates and pixel-aligned features from source images, the implicit network learns the occupancy, color, and gamma values for rendering the target image in novel views or poses.}
\label{fig:framework}
\end{figure*}

\section{Proposed Algorithm}

\subsection{Preliminary}

\subsubsection{Deformable NeRF}
For rendering an animatable human, we first define the canonical space where a person is in a standard pose that serves as a reference (see canonical pose in \Figref{fig:quali3}). 
Given the target space where the target pose is observed, we define a deep implicit function that predicts the occupancy probability and color at a 3D query point as
\begin{equation}
{G}(T(\mathbf{x}, \mathbf{p}), C(\mathbf{x}, \mathbf{p}))=(\alpha_{\mathbf{x}},{\mathbf c}_{\mathbf{x}}), 
\end{equation}
where $\alpha_{\mathbf{x}}$ and ${\mathbf c}_{\mathbf{x}}$ denote the occupancy probability and color, respectively, at the corresponding 3D coordinate $\bf x$ in the target space.
Occupancy represents the continuous probabilities of whether the space is filled or empty, and is directly used as an alpha value in the volume rendering process.
$T : (\mathbf{x}, \mathbf{p}) \mapsto \mathbf{x}^{c}$ denotes the skeletal deformation of the given pose $\mathbf p$, which maps a point $\mathbf{x}$ observed in target space where a person states in a target pose $\mathbf{p}$ to canonical space $\mathbf{x}^{c}$. 
Output values required for rendering target pixels are predicted from the network $G$ given a condition variable $C(\mathbf{x}, \mathbf{p})$, where the condition variable is defined as pixel-aligned features in the source space which will be described in the \secref{pose_aware}.


\subsubsection{Skeletal Deformation}

Following previous studies~\cite{peng2021animatable, weng2022humannerf, li2022tava}, we define the skeletal deformation by utilizing the linear blend skinning (LBS) function~\cite{weng2020vid2actor} to transform coordinates between different pose spaces. 
As the human skeleton consists of $K$ joints, the points in the target space are transformed to the canonical space using the inverse LBS function as follows,
\begin{equation}
\mathbf{x}^{c}=T(\mathbf{x}, \mathbf{p}) =\left(\sum_{k=1}^K w_k(\mathbf{x}) M^{\text{trg2can}}_k\right) \mathbf{x},
\end{equation}
where $w_k(\mathbf{x})$ is the blend weight for the $k$-th bone defined in the target space. 
$M^{\text{trg2can}}_k \in SE(3)$ is the transformation matrix of $k$-th skeleton part that maps the bone's coordinates from the target to the canonical space. Note that $M^{\text{trg2can}}$ can be computed from the given body pose $\mathbf{p}$. 

However, we cannot formulate the accurate skeletal deformation if identical blend weights are uniformly applied to various human identities that have different body shapes. 
To reconstruct the exact body shapes of various identities, we propose to solve for $w(\mathbf{x})$ using the weight field table, which will be described in detail in \secref{weight_field_table}. 

\subsection{Network Architecture}
\label{network_architecture}

The whole structure of the proposed model is shown in \figref{fig:framework}. 
The goal of our method is to generate target posed images from one or a few source images that contain unseen identity, poses, and views in the training. 

\subsubsection{Weight Field Table }
\label{weight_field_table}

To optimize the blend weight defined in the target space, the blend weight fields should be trained per frame included in the target motion sequences. This is significantly inefficient when we randomly change the motion sequence in the inference stage, because the additional blend weights are required to be optimized for every frame. 
Thus, we utilize the inverting equation between target and canonical space introduced in prior work~\cite{chen2021snarf, weng2020vid2actor, weng2022humannerf}, which enables approximating the blend weight of the target space using that of the canonical space. 
The $i$-th blend weight in the target space is derived as:
\begin{equation}
w_i(\mathbf{x})=\frac{w^c_i\left(M^{\text{trg2can}}_i(x)\right)}{\sum_{k=1}^K w^c_k\left(M^{\text{trg2can}}_k(x)\right)}.
\label{invert}
\end{equation}
It is known that this inverting equation using a single set of the canonical weight field yields better generalization than learning weight fields for every target space.


Since every human has a unique body shape, the weight field needs to be computed differently in order to transform the coordinates in the target space to the canonical space. 
We introduce {\it weight field table} that enables optimizing the distinct blend weights for diverse human identities.
We first allocate a table of learnable {\it shape codes} $\mathbf{S}\in\mathbb{R}^{L\times D}$, with each row representing a human identity. Here, $L$ denotes the number of identities in the training dataset and $D$ represents the dimension of the shape codes.
To reconstruct each subject, a shape code is sampled from the weight field table and decoded into the explicit volume representation using 3D convolution~\cite{weng2022humannerf}.
The output volume is then utilized as the canonical blend weight $w^c$ for the corresponding subject.


In the inference stage, we additionally define the new shape code for unseen identity, which can be optimized by leveraging the latent space that the weight field table learned. 
First, we initialize the shape code as the mean value of all shape codes in the weight field table.
Then, we optimize it through self-supervision by minimizing the Mean-Squared-Error (MSE) between the output images with the source poses and the given source images.
Note that the optimization process converges fast within short iterations.

\subsubsection{Pose-aware Pixel-aligned Features}
\label{pose_aware}

To render the target-posed images using an implicit function, the pixel-aligned feature \cite{yu2021pixelnerf, kwon2021neural, mihajlovic2022keypointnerf} is utilized as the condition variable $C(\mathbf{x}, \mathbf{p})$.
By encoding the source input images $\mathbf{I}$ using a convolutional encoder that has three downsampling layers, we extract the image feature ${\mathbf F}_{\mathbf{I}}$ by combining multi-scale intermediate features. 

To sample the pixel-aligned feature in a pose-aware manner, we transform the target coordinates to the source space using the blend weight defined in the target space as follows,
\begin{equation}
\mathbf{x}^{\text{src}}=\left(\sum_{k=1}^K w_k(\mathbf{x}) M^{\text{trg2src}}_k\right) \mathbf{x}.
\end{equation}

Then, by projecting the transformed source coordinate onto the image plane, we sample the pixel-aligned features $S({\mathbf F}_{I}, \pi(\mathbf{x}^{\text{src}}))$, where $\pi(\mathbf{x}^{\text{src}})$ denotes the 2D projection coordinate of $\mathbf{x}^{\text{src}}$ on the source image $\mathbf{I}$ and $S(\cdot)$ indicates the sampling function to sample the value at a query location using the bilinear interpolation.
Along with the the transformed canonical coordinates $\mathbf{x}^c$,  the sampled pixel-aligned features are used as inputs to the MLP network as a condition variable $C(\mathbf{x}, \mathbf{p})$ as follows, 

\begin{equation}
\begin{aligned}
C(\mathbf{x}, \mathbf{p} ) = \{S({\mathbf F}_{\mathbf{I}_n}, \pi({\mathbf{x}}^{\text{src}}_n))\}_{n=1 \cdots N},
\end{aligned}
\end{equation}
where $\mathbf{x}^{\text{src}}$ is the function of $\mathbf{x}$, and N is the number of source images to support the multiple image inputs. This will be described in detail in the following section.

\subsubsection{Multiview Stereo}

Multiple images provide more information about the person by removing geometric ambiguities of the one-shot case. 
To incorporate information from more views, we extend our model to take a random number of images as input.
We first sample the pixel-aligned feature from each view by transforming the target coordinate into each source space. 
The sampled features from all available views can be aggregated since they correspond to the same canonical location $\mathbf{x}^c$. 

Specifically, we decompose our implicit function $G$ into a feature embedding network $G_1$, an aggregating network $G_2$, and a weighting network $G_3$ for the feature fusion process. 
First, $G_1$ encodes each pixel-aligned feature into a latent feature embedding $\Phi_n$, where $n$ is the index of the input image among $N$ images. 
Instead of feeding the canonical coordinate directly into the implicit function, we extract voxel-aligned features based on the transformed canonical coordinates $\mathbf{x}^c$.
The voxel-aligned feature is acquired from a sparse 3D volume feature using SMPL vertices in the canonical space. 
We prepare sparse volume by dividing the 3D bounding box of canonical SMPL vertices with voxel size of 8mm, and extract the 3D volume feature using SparseConvNet~\cite{graham20183d}.  
Then, using the voxel-aligned feature at the transformed canonical coordinate $\mathbf{x}^c$, the aggregating network $G_2$ predicts the occupancy value $\alpha_{\mathbf{x}}$ and the intermediate feature.
The latent feature embeddings $\Phi_n$ are aggregated with a mean pooling operator, and then jointly blended with the intermediate feature to predict the color ${\mathbf c}^{\prime}_{\mathbf{x}}$.

In addition to the occupancy and color values, the weighting network $G_3$ predicts blending factor $\gamma_n$. 
$\gamma_n$ is used for blending the predicted color value with the ones sampled from the source images. 
Then, the final color value can be defined as follows, 
\begin{equation}
{\mathbf c}_{\mathbf{x}}=\sum_{n=1}^N \gamma_n \cdot S(\mathbf{I}_n, \pi({\mathbf{x}}^{\text{src}}_n))+\gamma^{\prime} \cdot {\mathbf c}^{\prime}_{\mathbf{x}}, 
\end{equation}
where $\gamma$ values are normalized with the softmax function.

\subsection{Training Objective} 

Based on the predicted occupancy and color value, we utilize volume rendering technique~\cite{mildenhall2021nerf} to synthesize novel views and poses of the given source images. 
The color of the target image is computed as follows, 
\begin{equation}
\mathbf{c}_{r}=\sum_{j=1}^{M} \alpha_{j} \prod_{k=1}^{j-1}\left(1-\alpha_{k}\right) \mathbf{c}_{j}, 
\end{equation}
where  ${\mathbf r}$ is the ray marched from the randomly selected target view and $M$ and $j$ are the number and the index of sampled points along a ray, respectively.

Our total training objective consists of following objectives: 1) the $\ell_{2}$ loss, 2) the perceptual loss, and 3) the opacity regularization.
We minimize the $\ell_{2}$ loss between the volume rendered color value and the ground-truth color value of the target-view image ${\mathbf I}^{\text{trg}}$. 
The reconstruction loss is formulated as
\begin{equation}
\mathcal{L}_{\ell_{1}}=\frac{1}{n_r}\sum_{i=1}^{n_r}| {\mathbf c}_{r}-{\mathbf c}^*_{r}|_{2},
\end{equation}
where $n_r$ is the number of rays, and ${\mathbf c}^*$ is the ground truth RGB value in the target-view image.

Also, we apply the perceptual loss to our training objective by minimizing the $\ell_{2}$ loss between the pre-trained VGG features of patches of generated target-view image and those of the ground-truth target-view image.
\begin{equation}
\mathcal{L}_{\text{vgg}}=\sum_{i=1}^{P} \left\|VGG\left({\mathbf p}_{i} \right)-VGG\left({\mathbf p}^*_{i} \right)\right\|_{2},
\end{equation}
where ${\mathbf p}$ and ${\mathbf p}^*$ is the synthesized image patches and the ground-truth image patches in the target view, respectively. $P$ is the number of image patches. 
We empirically found that utilizing high-level features of the pre-trained VGG network disturbs our network from reconstructing exact human body shapes, so we only employ the first shallow layer of the VGG network for the perceptual loss. 

Since our method targets the reconstruction of novel pose images from few-shot images, our method suffers from blurry image quality caused from depth ambiguity, which produces a non-zero occupancy value outside the object surface. 
To solve this problem, we impose a prior on the occupancy value $\alpha_{\mathbf{x}}$ to have zero value of entropy, which means the output occupancy value should be either 1 or 0 depending on the occupied spaces.

\begin{equation}
\mathcal{L}_\text{opacity}=\frac{1}{N}\sum_{i=1}^{N} \text{log}(\alpha_{\mathbf{x}_i})+\text{log}(1-\alpha_{\mathbf{x}_i}),
\end{equation}

where $\alpha_{\mathbf{x}}$ and $N$ are the output occupancy value and the number of the query points $\mathbf{x}$, respectively. 

Our full training objective functions for the network $G$ are written as 
\begin{equation}
\begin{aligned}
&\mathcal{L}_{\text{total}}=\mathcal{L}_{\text{vgg}}+\lambda_{\ell_{1}}\mathcal{L}_{\ell_{1}} +\lambda_{\text{opacity}}\mathcal{L}_{\text{opacity}},
\end{aligned}
\end{equation}
where $\lambda_{\ell_{1}}$ and $\lambda_{opacity}$ are hyperparameters determining the importance of each loss.

\begin{figure}[t]
\centering
\def\arraystretch{0.2}
\begin{tabular}{@{}c}
\includegraphics[width=1\linewidth]{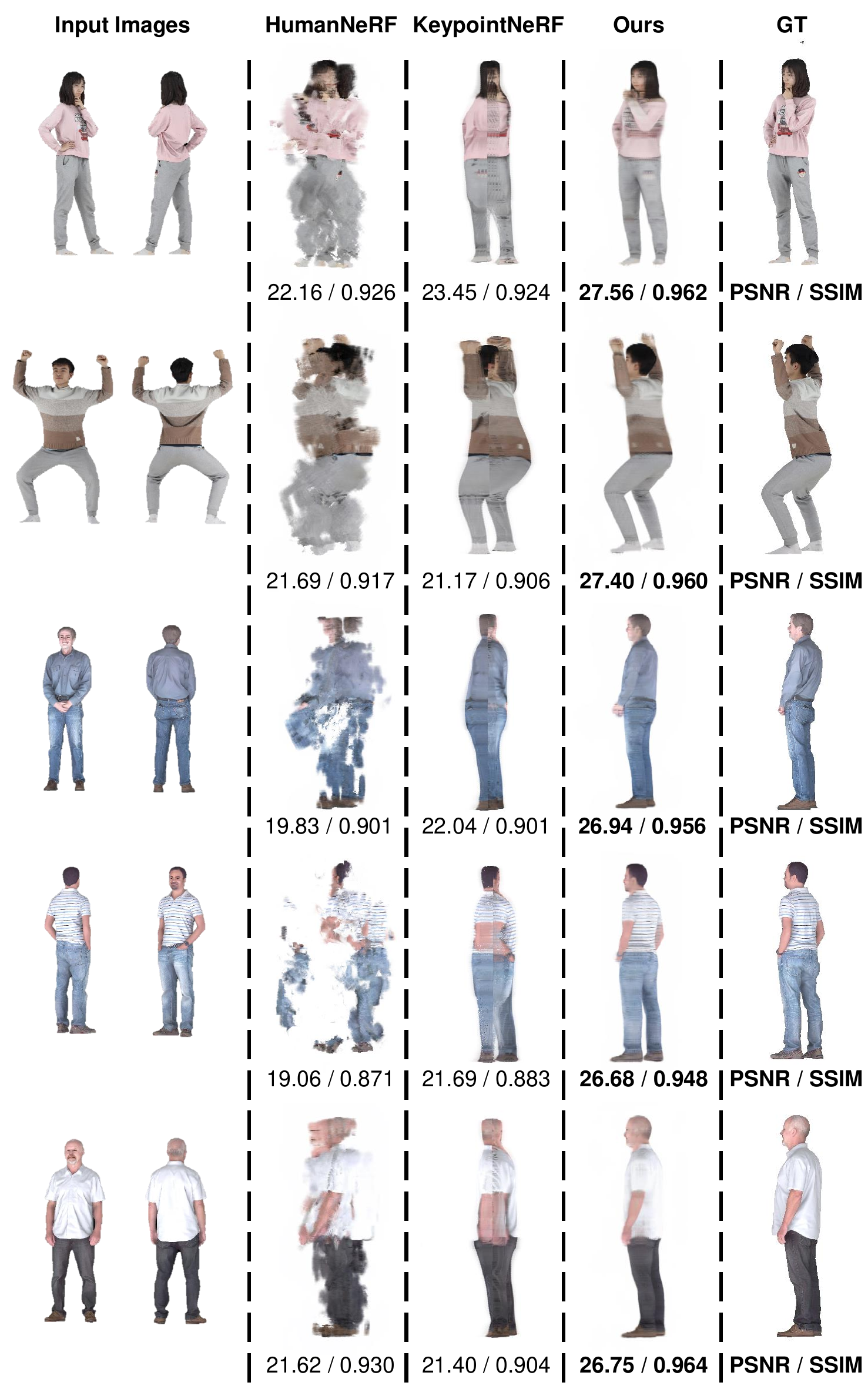} \\
\end{tabular}
\caption{Qualitative comparison of multiview synthesis on the THUman2.0 and Twindom test datasets. For each subject, given two source images that are sampled from 0 and 180 degrees, novel view images rendered from 90 degrees are presented in the next columns for each method.}
\label{fig:quali1}
\end{figure}

\section{Experimental Results}

\subsection{Dataset}

For training and evaluating our method, we collected 498 samples from the Twindom dataset and 526 samples from the THUman2.0~\cite{zheng2021deepmulticap} dataset. 
We rendered the textured human scans from 360 viewpoints at a resolution of 512$\times$512, and MuVS~\cite{huang2017towards} was adopted for preparing the ground-truth SMPL model. 
50 samples are selected from each dataset for evaluation of novel view synthesis and the remaining samples are used for training. 
For quantitative evaluation of novel pose synthesis, single view videos from ZJU-MoCap~\cite{peng2021neural} and Human3.6M~\cite{ionescu2013human3} are used, selecting the first camera of ZJU-MoCap and third camera of Human3.6M. 

\subsection{Training Details}
The network is trained with the learning rate $5e^{-4}$ using ADAM optimizer with $\beta_{1}=0.9$, $\beta_{2}=0.999$, except for the 3D CNN network that predicts the blend weight volume with the learning rate $5e^{-5}$. 
The network is trained for 1,000$K$ iterations with a batch size of 1, and 128 points are sampled per ray. 
For a source and target view image, the camera locations are randomly sampled among 360 degrees around the object, and 1-3 images are randomly selected for source images. 
In the training stage, we sample 5 patches with size $32 \times 32$ in each batch instead of casting random rays.
We use $\lambda_{\ell_{1}}=0.2$ and $\lambda_{\text{opacity}}=0.01$. The dimension of 256 is used for the shape codes of weight field table and volume size of 128 is used for the blend weight volume computed from the 3D convolution. 

\subsection{Performance Evaluations}

In this section, we compare the qualitative and quantitative results with other human rendering models to demonstrate that our method produces more realistic images in novel view and pose synthesis given few-shot images.
To the best of our knowledge, there exists no previous work that supports both generalizable novel view and novel pose synthesis when provided with few-shot images of unseen identities.
We select HumanNeRF~\cite{weng2022humannerf}, Ani-NeRF~\cite{peng2021animatable}, and KeypointNeRF~\cite{mihajlovic2022keypointnerf} as our baselines which are state-of-the-art human rendering methods. 
Note that HumanNeRF and Ani-NeRF are trained with source images given as inputs, and KeypointNeRF is retrained on our training dataset. 

\subsubsection{Novel View Synthesis}
\label{sec:exp_novelview}


Qualitative comparisons for novel view synthesis are shown in \Figref{fig:quali1}. 
Given two source images visualized in the first column, the images from novel view are visualized in the next columns for each method. Note that the images rendered from 0 and 180 degrees are used as source inputs and target images are rendered from 90 degrees. 
HumanNeRF~\cite{weng2022humannerf} has difficulty in reconstructing novel view due to depth ambiguity, because two images do not provide sufficient 3D information to train their model. 
KeypointNeRF, on the other hand, solve for depth ambiguity as an generalizable model, but it exhibits poor performance when fewer than three source images are provided as source inputs.
However, as noticeable in all examples, our method shows the best synthesis quality from any view, showing consistent pose and texture given in the source inputs. 
For each example, we measure PSNR and SSIM (structure similarity) for quantitative evaluation. Note that the mean value of multiview images rendered from 36 views spanning every 10 degrees in the horizontal axis is reported. 
We also compare our method with KeypointNeRF~\cite{mihajlovic2022keypointnerf} quantitatively for the test dataset in \tabref{tb:quanti}. 
Our method outperforms the baseline in overall. 
We notice that our method shows the robust performance when the number of the source images varies, while the performance of KeypointNeRF drops in large margin when the number of source images decreases.


\begin{table}[]
\centering
\renewcommand{\tabcolsep}{2mm}
\renewcommand{\arraystretch}{1.2}
\resizebox{1\linewidth}{!}{
\begin{tabular}{c|c|cc|cc}
\hline
\multirow{2}{*}{Methods}      & \multirow{2}{*}{\# of views} & \multicolumn{2}{c|}{THUman2.0}                       & \multicolumn{2}{c}{Twindom}                          \\ \cline{3-6} 
                              &                              & \multicolumn{1}{c|}{PSNR↑}          & SSIM↑          & \multicolumn{1}{c|}{PSNR↑}          & SSIM↑          \\ \hline
\multirow{2}{*}{KeypointNeRF} & 2 views                      & \multicolumn{1}{c|}{18.62}          & 0.903          & \multicolumn{1}{c|}{19.68}          & 0.890          \\ \cline{2-6} 
                              & 3 views                      & \multicolumn{1}{c|}{20.85}          & 0.931          & \multicolumn{1}{c|}{21.56}          & 0.920          \\ \hline
\multirow{2}{*}{Ours}         & 2 views                      & \multicolumn{1}{c|}{\textbf{23.67}} & \textbf{0.952} & \multicolumn{1}{c|}{\textbf{24.20}} & \textbf{0.948} \\ \cline{2-6} 
                              & 3 views                      & \multicolumn{1}{c|}{\textbf{24.47}} & \textbf{0.957} & \multicolumn{1}{c|}{\textbf{24.85}} & \textbf{0.952} \\ \hline
\end{tabular}
}
\vspace{1mm}
\caption{Quantitative comparison of multiview synthesis on the THUman2.0 and Twindom test datasets.}
\label{tb:quanti}
\end{table}

\begin{figure}[t]
\centering
\def\arraystretch{0.2}
\begin{tabular}{@{}c}
\includegraphics[width=1\linewidth]{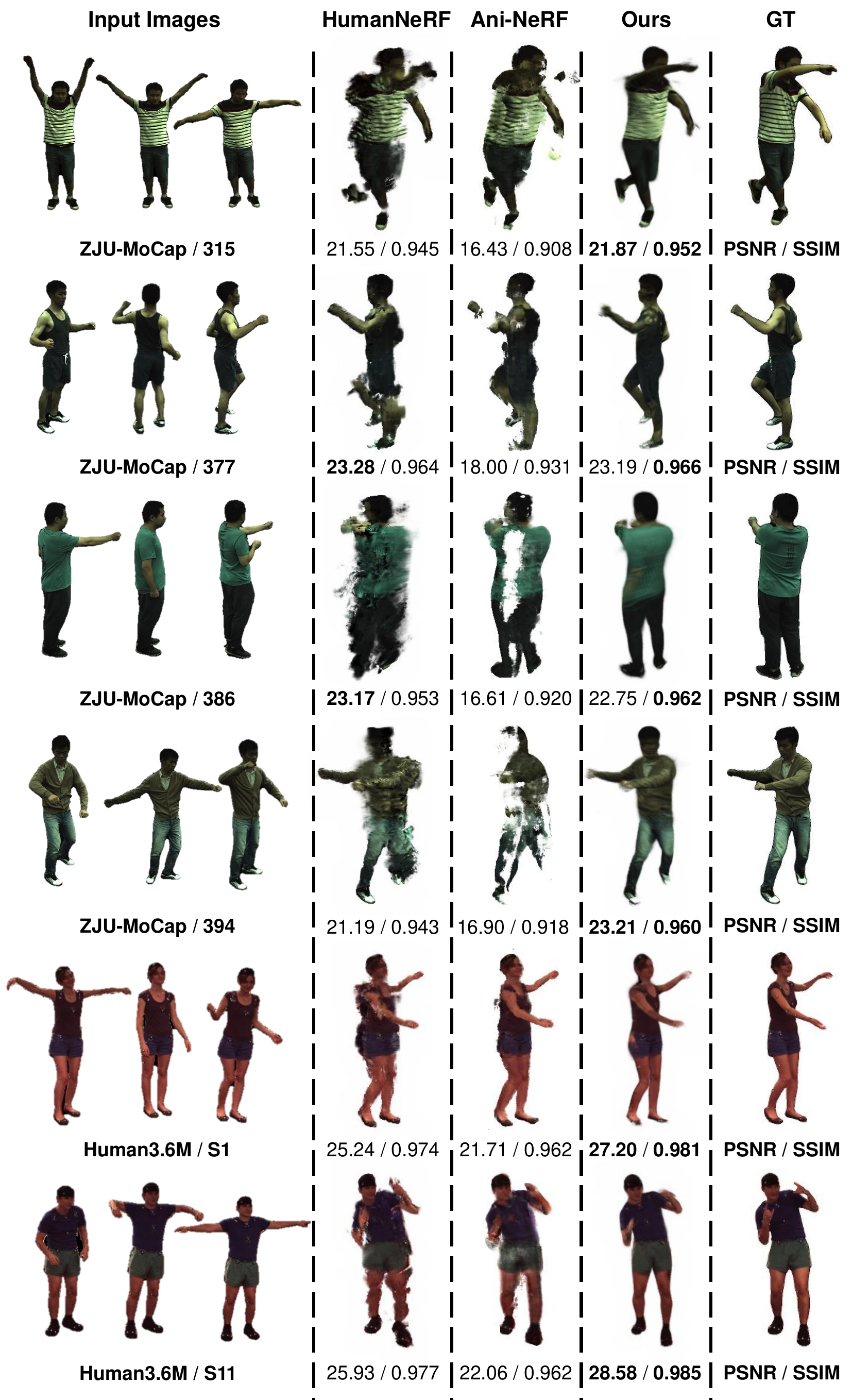} \\
\end{tabular}
\caption{Qualitative comparison of novel pose synthesis on the ZJU-MoCap and Human3.6M datasets. Given three source images, novel-posed images are presented for each method.}
\label{fig:quali2}
\end{figure}

\begin{figure*}[t]
\centering
\def\arraystretch{0.2}
\begin{tabular}{@{}c}
\includegraphics[width=0.89\linewidth]{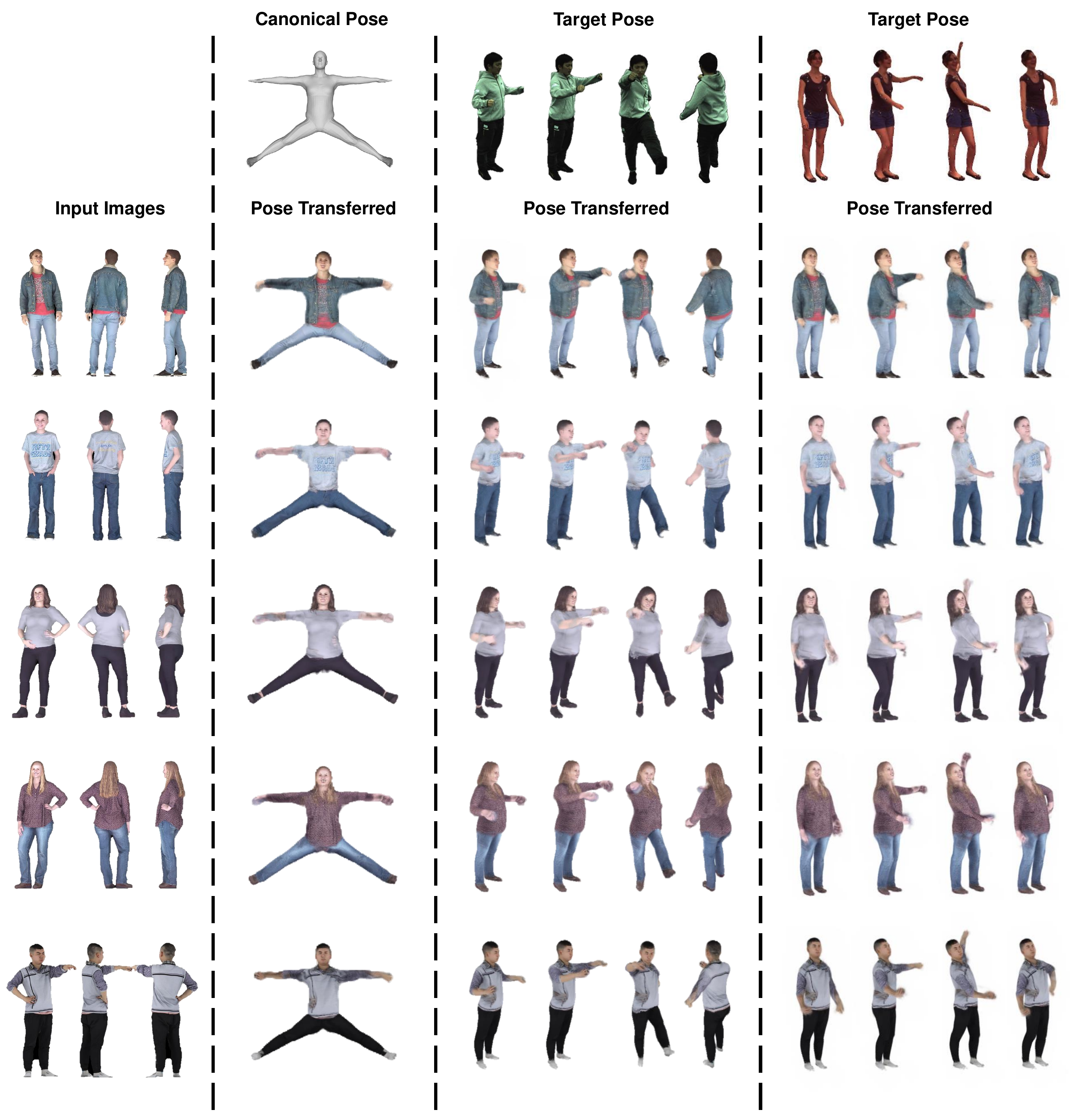} \\
\end{tabular}
\caption{Qualitative results of novel pose synthesis on the THUman2.0 and Twindom test dataset using the pose sequence in ZJU-MoCap and Human3.6M datasets. Given three source images, the canonical pose and novel-posed images for two different pose sequences are presented.}
\label{fig:quali3}
\end{figure*}

\subsubsection{Novel Pose Synthesis}

We present qualitative results for novel pose synthesis in \Figref{fig:quali2}. 
Given three source images presented in the first column, images with novel pose are visualized in the next columns for each method. 
We compare our method with HumanNeRF and Ani-NeRF that support novel pose synthesis. 
While both HumanNeRF and Ani-NeRF suffer from depth collapse when synthesizing novel pose images, \ourmodel~ successfully renders novel pose images that closely resemble ground-truth images.
For a few cases, the baseline method was only able to synthesize a novel pose with minimal depth collapse when the target pose was very close to the source pose (see S1 of Human3.6M).
In addition, while Ani-NeRF requires additional optimization process for every frame, our method directly generates novel pose images without additional optimization when a new pose sequence is given. 
For each example, we report PSNR and SSIM measuring multi-pose images rendered for the first 40 frames in the given video, with intervals of 10 frames for the ZJU-MoCap dataset and 5 frames for the Human3.6M dataset.
Our method outperforms the baselines in most cases. 
While HumanNeRF may exhibit higher PSNR scores on samples 377 and 386 in ZJU-MoCap, it is important to note that this is likely due to the test pose sequences being similar to the source poses.
In contrast, our method is able to generalize to any novel pose without a decrease in performance, and shows robust performance in terms of SSIM score for all the examples. 
Note that all output images are rendered in white background to observe the depth collapse.
We also present various outputs of novel pose synthesis in \Figref{fig:quali3} to demonstrate that our model can successfully render a human body in a novel pose when various pose sequences are given.

\subsection{Ablation Studies}


\subsubsection{Weight Field Table}

\begin{figure}[t]
\centering
\def\arraystretch{0.2}
\begin{tabular}{@{}c}
\includegraphics[width=0.9\linewidth]{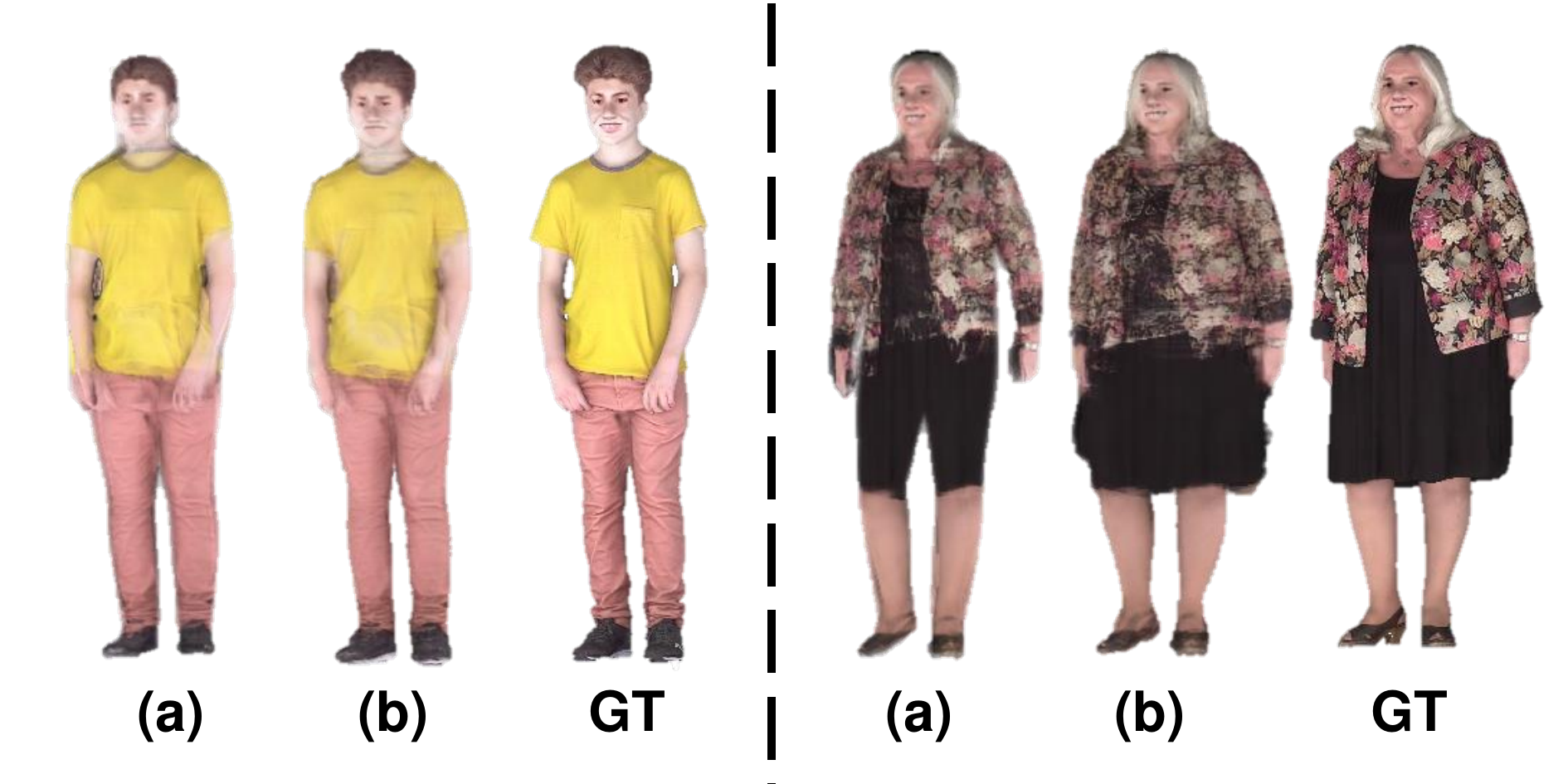} \\
\end{tabular}
\caption{\small Ablation study of weight field table. (a) denotes the output image of the model trained without the weight field table, and (b) denotes that of our full model.}
\label{fig:abl1}
\end{figure}

Here, we demonstrate the effectiveness of our weight field table both quantitatively and qualitatively.
Since the weight field determines the blend weight for each joint in the human body, it actually determines the translation for each query point in the target space when transformed into the canonical space. 
Thus, it generates a different body shape for each identity by predicting varying occupancy values at the same query point in the target space. 
As shown in \Figref{fig:abl1}, the model trained without the weight field table fails to reconstruct the exact shape of the source human body. In contrast, our full model successfully reconstructs the complex human body shape, including features such as hair and a voluminous skirt. 
As reported in the quantitative result in \tabref{tb:abl2}, following the same protocol for evaluating multiview sythesis in \secref{sec:exp_novelview}, our full model achieves higher reconstruction accuracy compared to the model trained without the weight field table. 
This indicates the weight field table helps in reconstructing the exact shape of a human body.

\begin{table}[]
\centering
\renewcommand{\tabcolsep}{2mm}
\renewcommand{\arraystretch}{1.2}
\resizebox{1\linewidth}{!}{
\begin{tabular}{c|cc|cc|cc}
\hline
\multirow{2}{*}{Dataset} & \multicolumn{2}{c|}{\begin{tabular}[c]{@{}c@{}}Without \\ Weight Field Table\end{tabular}} & \multicolumn{2}{c|}{\begin{tabular}[c]{@{}c@{}}100-step\\ Optimization\end{tabular}} & \multicolumn{2}{c}{\begin{tabular}[c]{@{}c@{}}200-step\\ Optimization\end{tabular}} \\ \cline{2-7} 
                         & \multicolumn{1}{c|}{PSNR↑}                             & SSIM↑                             & \multicolumn{1}{c|}{PSNR↑}                          & SSIM↑                          & \multicolumn{1}{c|}{PSNR↑}                          & SSIM↑                         \\ \hline
THUman2.0                & \multicolumn{1}{c|}{21.75}                             & 0.939                             & \multicolumn{1}{c|}{23.98}                          & 0.955                          & \multicolumn{1}{c|}{\textbf{24.47}}                          & \textbf{0.957}                         \\ \hline
Twindom                  & \multicolumn{1}{c|}{22.13}                             & 0.937                             & \multicolumn{1}{c|}{24.51}                          & 0.951                          & \multicolumn{1}{c|}{\textbf{24.85}}                          & \textbf{0.952}                         \\ \hline
\end{tabular}
}
\vspace{1mm}
\caption{Ablation study of weight field table and shape code optimization. Note that we quantitatively measured multiview synthesis on the THUman2.0 and Twindom test dataset using three views as source inputs. }
\vspace{-3mm}
\label{tb:abl2}
\end{table}

\subsubsection{Shape Code Optimization}
Thanks to our proposed weight field table which forms the latent space for various human body shapes, we are able to reconstruct the body shape of unseen identities by optimizing the new shape codes in the inference stage. 
We assert that the optimization process is quick to converge, as it takes only 129.7 seconds for 200 iterations to reach convergence. 
To further demonstrate the efficiency of the optimization process,
we have included the results of both 100-step and 200-step optimization in \tabref{tb:abl2}.
The optimization process is measured on a machine equipped with an AMD EPYC 7502 CPU and an NVIDIA RTX 3090 GPU.

\subsubsection{Limitations}
\begin{figure}[t]
\centering
\def\arraystretch{0.2}
\begin{tabular}{@{}c}
\includegraphics[width=0.9\linewidth]{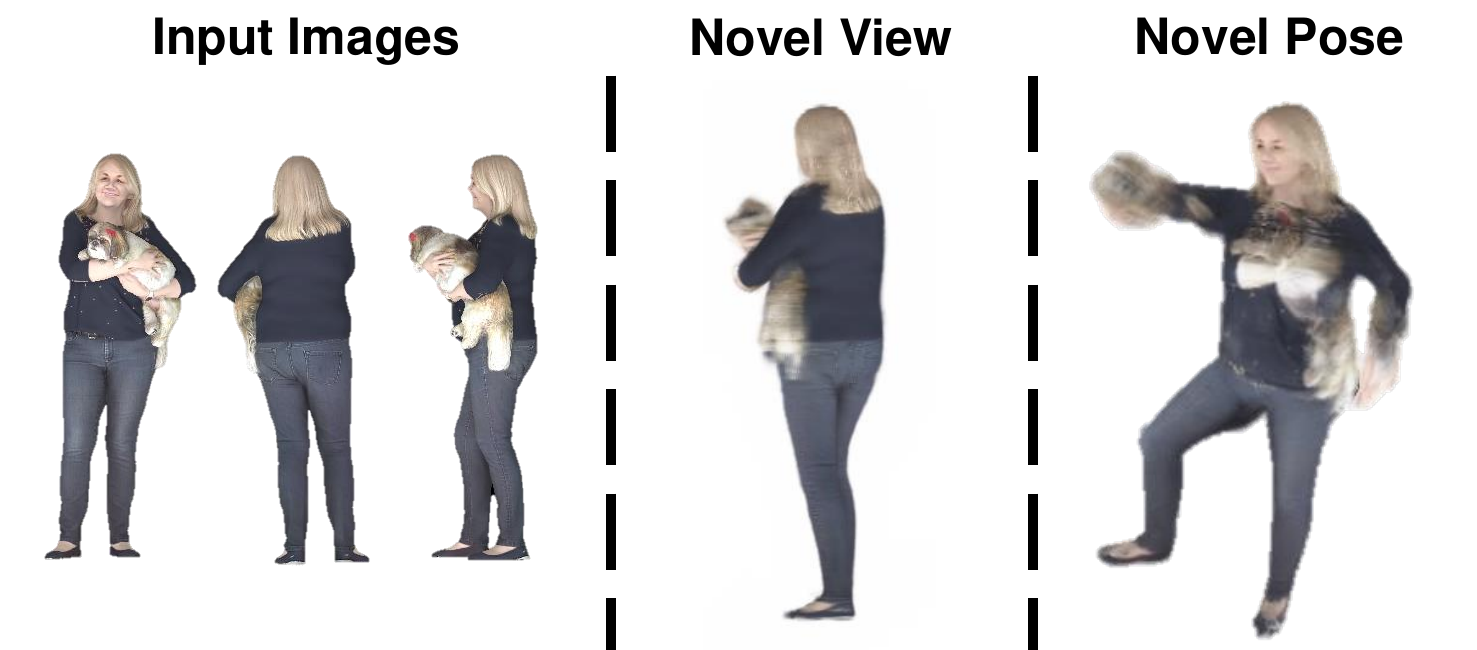} \\
\end{tabular}
\caption{\small Failure cases. We present the output images rendered in a novel view and pose from the given three source images.}
\label{fig:abl2}
\end{figure}

We present failure cases of our proposed method in this section.
Since our method is designed to learn the prior knowledge of the human body, it is not intended to handle other objects such as animals or non-living items.
As presented in \Figref{fig:abl2}, our method fails to render realistic images in a novel pose when a non-human object is included in the source images. 
In addition, our method encounters difficulties in cases of extreme self-occlusion caused by challenging poses, such as a sitting position, or by the presence of additional objects.
We plan to address these issues in our future work to improve the robustness of our method in challenging cases.

\section{Conclusion}

In this paper, we have introduced a novel human rendering model~\ourmodel, which generates animatable human scenes from an unseen human identity and poses conditioned on a single or a few images of the person. 
To the best of our knowledge, the proposed algorithm is the first generalizable method that synthesizes both novel view and pose images from few-shot inputs. 
Thanks to the {\it weight field table} that learns the unique deformations fields of various human identities and the pose-aware pixel-aligned feature in the source space, our method is able to reconstruct the accurate body shape and texture from the given source images. 
Extensive experiments demonstrate that our method achieves state-of-the-art performance quantitatively and qualitatively in multiview and novel pose synthesis from few-shot images.

{\small
\bibliographystyle{ieee_fullname}
\bibliography{egbib}
}

\end{document}